%% file: main.tex
\documentclass[10pt,twocolumn,letterpaper]{article}

\usepackage[pagenumbers]{cvpr} 


\definecolor{cvprblue}{rgb}{0.21,0.49,0.74}
\usepackage[pagebackref,breaklinks,colorlinks,allcolors=cvprblue]{hyperref}
\usepackage{amsmath}
\usepackage{amssymb}
\usepackage{booktabs}
\usepackage{multirow} 
\usepackage{float}

\def\paperID{xxxx} 
\def\confName{CVPR}
\def\confYear{2025}

\title{
    \begin{minipage}[r]{0.1\textwidth} %
        \centering
        \includegraphics[height=1.5cm]{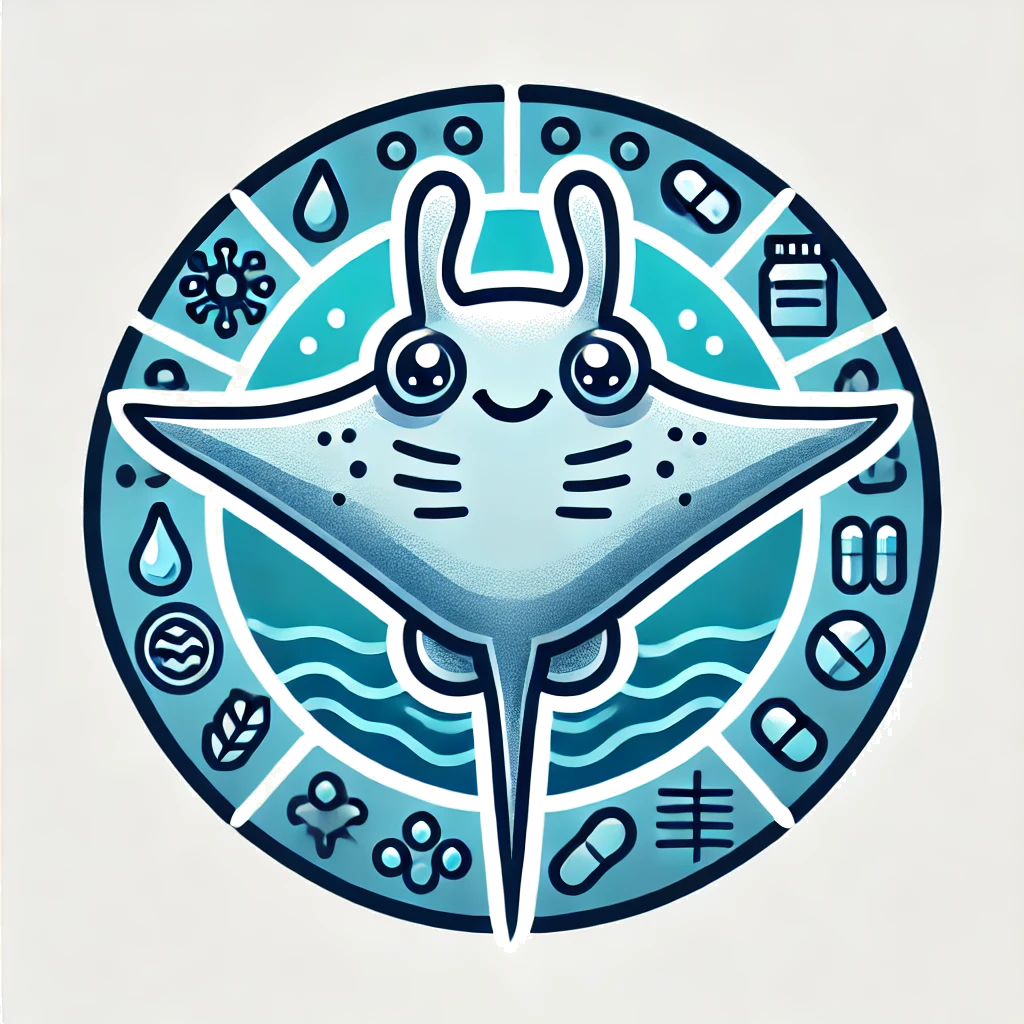} %
    \end{minipage}%
    \hspace{-0.06\textwidth} %
    \begin{minipage}[l]{0.88\textwidth}
        \centering
        MANTA: A Large-Scale Multi-View and Visual-Text Anomaly Detection Dataset for Tiny Objects \\
    \end{minipage}
\vspace{-1.8em}
}

\author{%
	Lei Fan$^{1*}$ \thanks{Corresponding author: \tt lei.fan1@unsw.edu.au}   \hspace{2em}  Dongdong Fan$^{2}$ \hspace{2em}  Zhiguang Hu$^{3}$ \hspace{2em} Yiwen Ding$^{2}$ \\ Donglin Di$^{4}$ \hspace{2em}  Kai Yi$^{5}$ \hspace{2em} Maurice Pagnucco$^{1}$ \hspace{2em} Yang Song$^{1}$ \\ 
	\vspace{-1em}
    \\
	$^{1}$UNSW Sydney \hspace{2em} $^{2}$Gaozhe \hspace{2em} $^{3}$SCAU \hspace{2em} $^{4}$Li Auto \hspace{2em} $^{5}$University of Cambridge\\ 
	\tt \small \url{https://grainnet.github.io/MANTA}
    \vspace{-1em}
}

\def\eg{\emph{e.g., }}

\let\oldtwocolumn\twocolumn
\renewcommand\twocolumn[1][]{%
    \oldtwocolumn[{#1}{
    \vspace{-3.5em}
    \begin{center}
           \includegraphics[width=.98\textwidth]{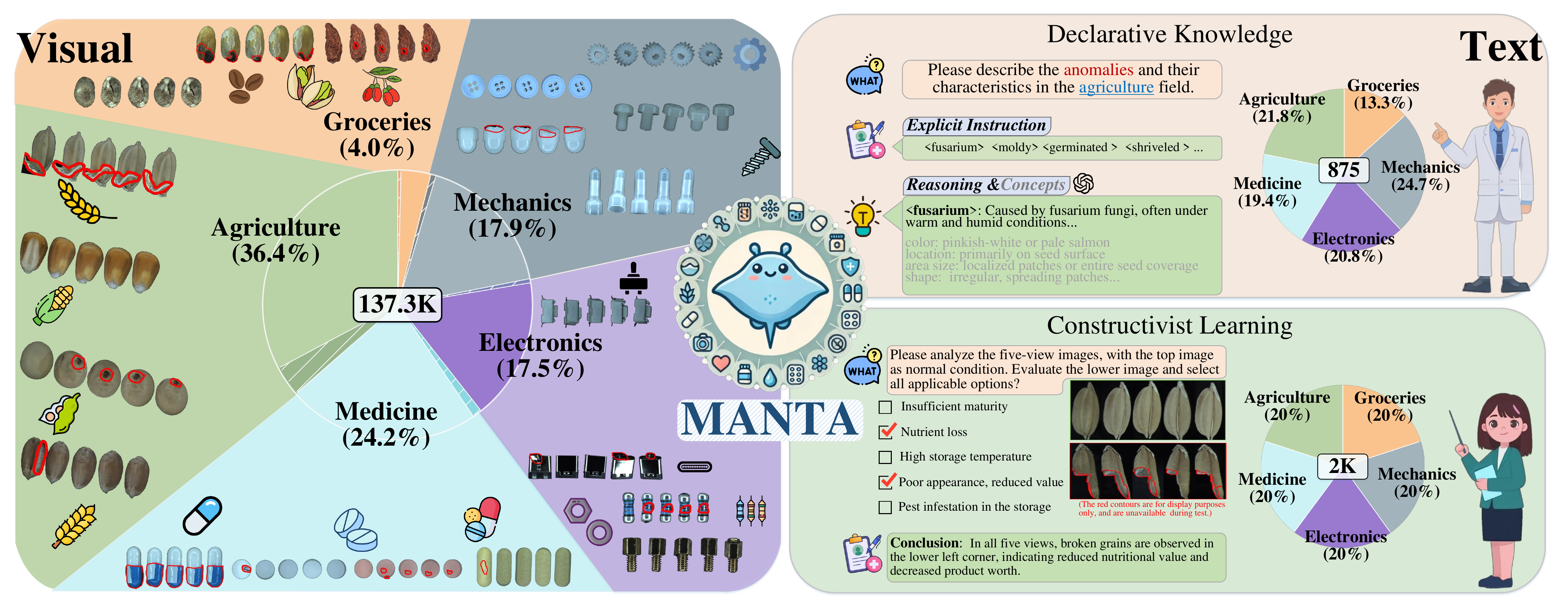}    
           \captionof{figure}{\textbf{Overview of MANTA}. It consists of both visual and text components. The visual part includes over 137K multi-view images spanning five domains. The text part is divided into two subsets: \textit{Declarative Knowledge}, comprising 875 words describing common anomalies, and \textit{Constructivist Learning}, which includes 2K Image-text multiple-choice questions.}           
           \label{fig:poster}
        \end{center}
    }]
}

\begin{document}
\maketitle

\renewcommand{\abstractname}{\vspace{-3em}\bfseries Abstract}

\begin{abstract}
  We present \textbf{MANTA}, a visual-text anomaly detection dataset for tiny objects. The visual component comprises over 137.3$K$ images across 38 object categories spanning five typical domains, of which 8.6$K$ images are labeled as anomalous with pixel-level annotations. Each image is captured from five distinct viewpoints to ensure comprehensive object coverage. The text component consists of two subsets: \textit{Declarative Knowledge}, including 875 words that describe common anomalies across various domains and specific categories, with detailed explanations for $\left< what, why, how \right>$,  including causes and visual characteristics; and \textit{Constructivist Learning}, providing 2$K$ multiple-choice questions with varying levels of difficulty, each paired with images and corresponded answer explanations. We also propose a baseline for visual-text tasks and conduct extensive benchmarking experiments to evaluate advanced methods across different settings, highlighting the challenges and efficacy of our dataset.

\end{abstract}

\input{chapters/intro}

\input{chapters/related}

\input{chapters/MVAD}

\input{chapters/method}

\input{chapters/exp}

\input{chapters/conc}

\clearpage
{
    \small
    \bibliographystyle{ieeenat_fullname}
    \bibliography{main}
}


\end{document}

%% file: chapters/intro.tex
\section{Introduction}

Anomaly detection (AD) has gained significant attention across various fields \cite{roth2022towards,liu2024deep,han2022adbench,schmidl2022anomaly,su2021hyperspectral,fan2022grainspace}. High-quality datasets, \eg MVTec \cite{mvtec_ad}, play a crucial role in advancing AD models and establishing unified benchmarks. These datasets are predominantly designed for industrial objects of large physical sizes and typically rely on single-view cameras to capture individual images, where anomalies are often more discernible and easier to detect.

In real-world applications, particularly those involving natural objects and manufactured items, there is a substantial volume of samples with tiny physical sizes \cite{liu2021survey}. Unlike larger objects that can be readily inspected \cite{tong2022deep}, identifying anomalies in these tiny objects requires considerable labor and cost, making it a critical yet underexplored area. For example, inspecting medical pills, agricultural seeds, or mechanical components presents unique challenges. Traditional single-view imaging setups \cite{liu2024deep} often fail to capture critical perspectives due to inherent design limitations, resulting in incomplete visual information. Compared to larger objects with clear textures and structural characteristics \cite{hojjati2024self}, tiny objects require high concentration during data collection and inspection.

In this paper, we present \textbf{MANTA}, a \textbf{M}ulti-view \textbf{AN}omaly detection with \textbf{T}ext \textbf{A}nalysis dataset for tiny objects, as illustrated in Figure~\ref{fig:poster}. We acquired a large number of tiny objects from common domains in daily life and industrial environments, and designed a prototype equipped with five high-resolution cameras to capture visual data from five different angles, ensuring comprehensive surface coverage for each object. We then provide high-quality data with detailed annotations for both visual and textual components. Specifically, the visual component comprises over 137.3$K$ multi-view images, including over $8K$ images annotated as anomalies with pixel-level annotations. Each multi-view image captures five viewpoints of a single tiny object. These images cover 38 categories, \eg wheat kernels, medical tablets, Type-C interfaces, and screws, grouped into five typical domains: agriculture, medicine, electronics, mechanics, and groceries.

The text component consists of two subsets: Declarative Knowledge (\textit{DeclK}) and Constructivist Learning (\textit{ConsL}). Specifically, \textit{DeclK} summarizes expert domain knowledge through text descriptions at three levels including explicit instruction, reasoning and concepts corresponding to \textit{$<$what, why, how$>$}. We provide 875 instruction words covering both domain-specific and category-specific anomalies across five domains. Each instruction is represented by a noun or adjective to describe types of anomalies (\eg scratches); reasoning explains the cause of the anomaly (\eg physical impact); and concepts outline common visual characteristics of the anomaly, \eg typical colors like white and gray, locations often found on the surface, irregular sizes, and common shapes appearing as lines. \textit{ConsL} is designed to emulate a human learning approach through visual comparisons and corresponding multiple-choice questions (MCQ), reflecting the process of human learning by analyzing images and selecting answers based on visual cues. We provide over 2000 MCQs with varying levels of difficulty, each consisting of a pair of normal-normal or normal-anomaly images, accompanied by five domain-specific options, answers, and explanations. 

We also propose a baseline by fine-tuning a pretrained multimodal language-image model \cite{li2023blip} to perform our \textit{ConsL} task. Furthermore, we conduct extensive experiments to evaluate advanced methods on our dataset under various settings, \eg one-for-one, multi-class, and text prompt learning. Our contributions are listed as follows:

\begin{itemize}
    \item  We release a large-scale visual-text AD dataset: MANTA. It contains over 137$K$ multi-view images of tiny objects spanning 38 categories across five domains.
    \item MANTA includes two text subsets: \textit{DeclK}, summarizing expert knowledge for various anomalies at three levels, and \textit{ConsL}, providing multiple-choice questions paired with images and explanations. 
    \item We propose a baseline model by fine-tuning foundation model to perform visual-language tasks on \textit{ConsL}.
    \item We conduct comprehensive benchmarking experiments, involving over 400 evaluations across five different settings on our dataset, demonstrating the challenges and efficacy of MANTA.
\end{itemize}

%% file: chapters/related.tex
\section{Related Work}

\subsection{Anomaly Detection Datasets}
Early studies \cite{tabernik2020segmentation,carrera2016defect,mery2015gdxray,btad_ad,jezek2021deep,huang2020surface,fan2023identifying} focused on specific domains and provided limited categories for visual discrimination of structural defects. Datasets like MVTec AD \cite{mvtec_ad}, VisA \cite{visa_ad} and MVTec LOCO AD \cite{bergmann2022beyond} introduce a diverse range of objects and textures for standard benchmarks. Although these datasets offer high-quality images and annotations, they are constrained to single-view setups and lack standardized backgrounds. MVTec 3D \cite{visapp22} and Real3d-ad \cite{liu2024real3d} provide point cloud data acquired by 3D sensors, but are constrained by limited sample volumes. Recently, Real-IAD \cite{real_ad} contains diverse objects with multi-view images, though full-surface inspection requires manual intervention. 

In contrast to these studies, which rely on single-view or incomplete multi-view setups, we provided a dataset for tiny objects captured from five views, ensuring comprehensive coverage of their full surfaces.

\begin{figure*}[h]
	\centering
	\includegraphics[width=1\textwidth]{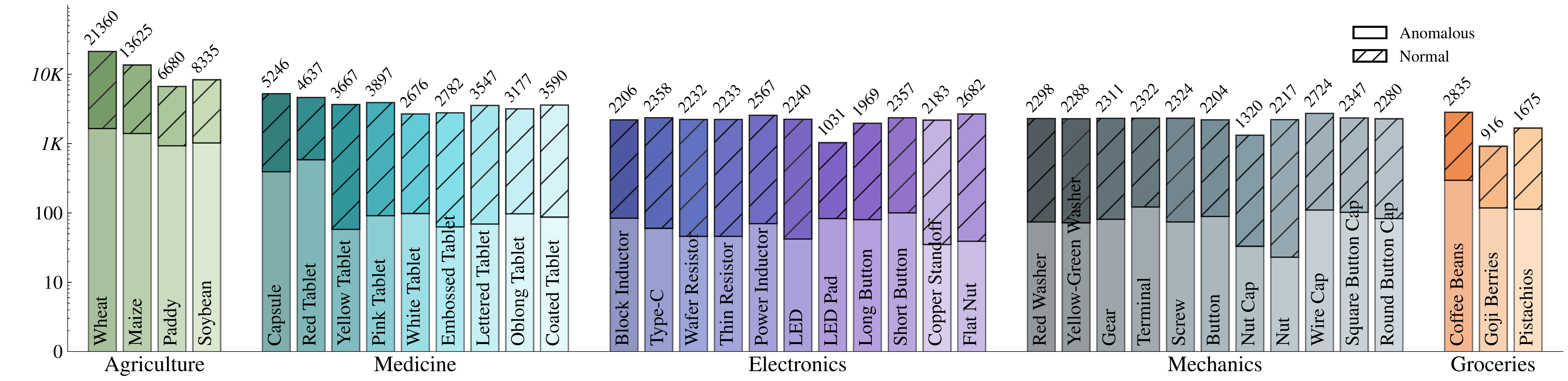}
	\caption{\textbf{The data distribution of the visual component in MANTA.} It contains 137,338 multi-view images across 38 categories from five typical domains: Agriculture, medicine, electronics, mechanics, and groceries. Each histogram represents a category, with different color groups indicating domains. For better visualization, a $\log_{10}$ scale is used on the y-axis.}     
	\label{fig:data_dis}
\end{figure*}

\subsection{Anomaly Detection Settings}
Visual anomaly detection task is typically formulated as an unsupervised manner \cite{pang2021deep}, where only normal samples are available for model training. 
Existing studies can be broadly categorized into reconstruction-based \cite{deng2022anomaly,liu2023diversity,ristea2022self}, synthesis-based \cite{zhang2024realnet,li2021cutpaste,zavrtanik2021draem,liu2023simplenet}, embedding-based \cite{defard2021padim,ruff2018deep}, and memory-based methods \cite{patchcore,bae2023pni}. These methods effectively transform the unsupervised setting into a supervised one by constructing pretext tasks or learning compact representations of normal samples. Unlike these studies, some studies focused on developing `one-for-all' models \cite{uniAD,he2024diffusion,guo2023template} capable of handling multiple categories simultaneously. Recently, several studies \cite{sato2023prompt,jeong2023winclip,liu2024unsupervised,zhu2024toward,huang2024adapting,li2024promptad} explored text prompts, such as defect names, to leverage the capabilities of foundation models like CLIP \cite{radford2021learning}.

In addition to the above settings, MANTA's multi-view setup supports both view-level and object-level AD evaluations. Moreover, two text subsets cover domain knowledge and multiple-choice questions, facilitating comprehensive analysis of visual-language models.

%% file: chapters/MVAD.tex
\section{MANTA}

We begin with data acquisition, including data collection, data-capturing prototype, and preprocessing procedure. Next, we introduce MANTA, detailing its visual and textual components and the relevant challenges.

\subsection{Data Acquisition}

\textbf{Sample Collection}. We gathered and curated a large-scale repository comprising over 300$K$ tiny objects across various domains including agriculture, pharmaceuticals, industry, food, to daily items. The physical sizes of these objects mainly range from $4^3$ to $20^3 mm^3$. We preprocessed and filtered approximately 200$K$ objects for subsequent data construction. To ensure data balance and increase the number of anomalous samples, we manually introduced a series of controlled defects. Detailed information regarding object categories and physical size can be found in the \textit{Supp}.

\begin{figure}[t]
	\centering
	\includegraphics[width=0.48\textwidth]{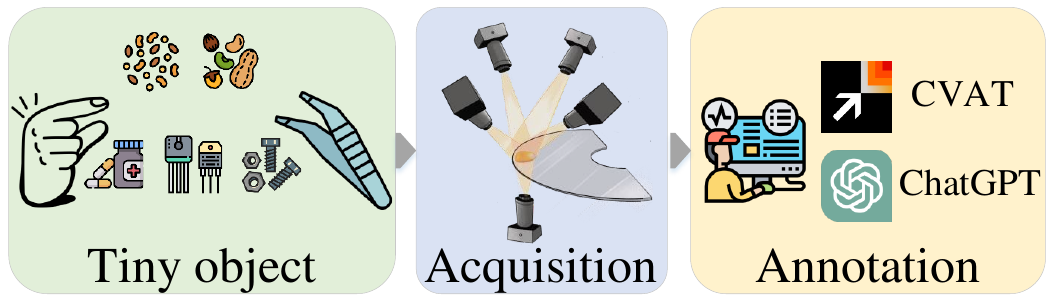}
	\caption{\textbf{Data acquisition}. We collected tiny objects from various domains, constructed a prototype to capture visual information, and annotated them using CVAT and ChatGPT.} 
	\label{fig:data-collection}
\end{figure}

\textbf{Prototype}. We constructed a prototype equipped with five high-resolution cameras to capture the surface information of tiny objects individually, as shown in Figure ~\ref{fig:data-collection}. The tiny object is placed on a transparent plate, with four cameras arranged in a quadrilateral formation, tilted downward at 30$^\circ$, and one additional camera positioned vertically beneath, pointing upward. Given the small size of tiny objects, achieving high-quality, low-noise images is crucial. To ensure this, shadowless light sources are placed on both sides, providing consistent and uniform illumination. This setup captures five-view images for each object, where each camera operates at 1170 DPI (the bottom one operates at 1250 DPI), ensuring high-quality images capable of detecting defects as small as $0.2^3mm^3$.

\subsection{Image Annotations and Statistics}

\textbf{Preprocessing and Annotations}. The raw image from each camera has a resolution of $1272\times1016$. We used a mechanical setup to separate tiny objects individually. After capturing visual information, we applied basic morphological operations to coarsely localize the object's center, followed by cropping to a fixed size to ensure consistent scaling across similar objects while preserving relative size information. Background artifacts were removed manually, and image quality was calibrated using a standardized color plate to enhance uniformity and consistency across the dataset. For each object, the five processed images were aligned and vertically merged into a single five-view composite image based on the same camera order.

We recruited a team of four expert inspectors to classify images into normal or anomalous categories, with pixel-level annotations highlighting the anomalous regions using CVAT \cite{CVAT_2024}. Additionally, two lightweight classification \cite{he2016deep} and segmentation models \cite{ronneberger2015u} were trained to filter out low-quality or annotation errors through a progressive holdout process. All results were then manually re-validated to ensure accuracy and consistency.

\begin{table}[b]
\caption{Comparison with various AD datasets. \textit{\#Cate} and \textit{\#O/Poly} refer to the number of object categories and labeled polygons per object respectively. The \textit{object} row indicates that each sample represents multiple views of a single object.}
 \centering
\resizebox{0.5\textwidth}{!}{
    \begin{tabular}{llllcccc}
    \toprule
    \multirow{2}[0]{*}{Datasets('year)}   & \multicolumn{3}{c}{Sample Statistics} & \multicolumn{2}{c}{Attributes} & \multicolumn{2}{c}{Annotations}  \\
    \cmidrule(lr){2-4}  \cmidrule(lr){5-6} \cmidrule(lr){7-8} 
      & \textit{\#Norm} & \textit{\#Anom} & \textit{\#Tota} & \textit{\#Cate} & View & \textit{\#O/Poly} & Modal \\
    \cmidrule(lr){1-8}
    BTAD'21 \cite{btad_ad}    & 952  & 290  & 1,242 & 3       & \textit{Single}     & 2.6  & \includegraphics[width=0.3cm]{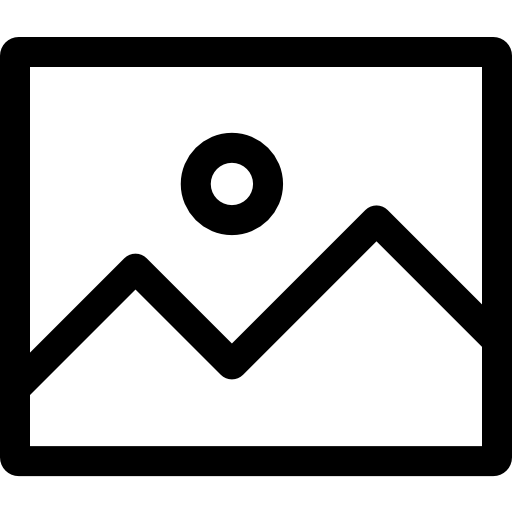} \\
    MPDD'21 \cite{jezek2021deep}   & 1,064  & 282  & 1,346 & 6       & \textit{Single}     & 1.8  & \includegraphics[width=0.3cm]{imgs/img-icon.png} \\
     MVTec AD'21  \cite{mvtec_ad}  & 4,096  & 1,258  & 5,354  & 15      & \textit{Single}     & 1.5  & \includegraphics[width=0.3cm]{imgs/img-icon.png} \\
      VISA'22 \cite{visa_ad}  & 9,621  & 1,200  & 10,821 & 12     & \textit{Single}    & 3.5  & \includegraphics[width=0.3cm]{imgs/img-icon.png} \\
     LOCO-AD'22 \cite{bergmann2022beyond}  & 2,347  & 993  & 5,354   & 5    & \textit{Single}    & 1.2  & \includegraphics[width=0.3cm]{imgs/img-icon.png} \\
   GoodsAD'24 \cite{zhang2024pku}    & 4,464  & 1,659  & 6,123 & 6       & \textit{Single}     & 1.1  & \includegraphics[width=0.3cm]{imgs/img-icon.png} \\
     Read-IAD'24 \cite{real_ad}   & 99,721  & 51,329  & 151,050   & \multirow{2}[0]{*}{30}     & \textit{Single}    & \multirow{2}[0]{*}{3.7}  & \multirow{2}[0]{*}{\includegraphics[width=0.3cm]{imgs/img-icon.png}} \\
      Read-IAD (\textit{object})   & 14,568  & 15,642  & 30,210   &      & \textit{Multi}    & &  \\
     \cmidrule(lr){1-8}
    MANTA (\textit{single})  & 652,455  & 34,235  & 686,690   & \multirow{2}[0]{*}{38}      & \textit{Single}     & \multirow{2}[0]{*}{5.1}  & \multirow{2}[0]{*}{ \includegraphics[width=0.3cm]{imgs/img-icon.png} \includegraphics[width=0.32cm]{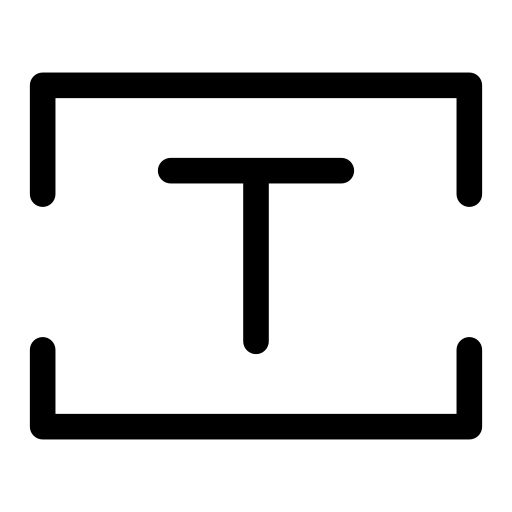}} \\
     MANTA (\textit{object})  & 128,721  & 8,617 & 137,338   &      & \textit{Multi}    &   &  \\
     \bottomrule
    \end{tabular}%
}
\label{tab:datasets}
\end{table}

\textbf{Data Statistics}. The visual component in MANTA contains 137,338 multi-view images. It includes 38 categories spanning five typical domains, as shown in Figure \ref{fig:data_dis}. The ontology and examples can be found in the \textit{Supp}.  All images are in RGB format, with resolutions ranging from $162\times705$ to $1117\times3365$ pixels. Among these, 8,617 images are categorized as anomalous and include mask annotations to highlight the anomalous regions. Considering the imbalance between normal and anomalous data, for each category with $N$ anomalous images, we partitioned the training and testing sets by randomly selecting $2N$ normal images along with all anomalous images to form the test set, while the remaining normal data was used for the training set. Compared to existing AD datasets, as shown in Table~\ref{tab:datasets}, MANTA leads significantly in both dataset size, category, and annotation volume. Notably, similar to the multi-view Real-IAD dataset, we report both \textit{single}-view (each view treated as a separate sample) and \textit{object}-view (multi-view images of an object treated as a single sample) information. Additionally, we further provide text descriptions in the following section. Additionally, we provide text descriptions in the subsequent section.

\begin{figure}[b]
	\centering
	\includegraphics[width=0.48\textwidth]{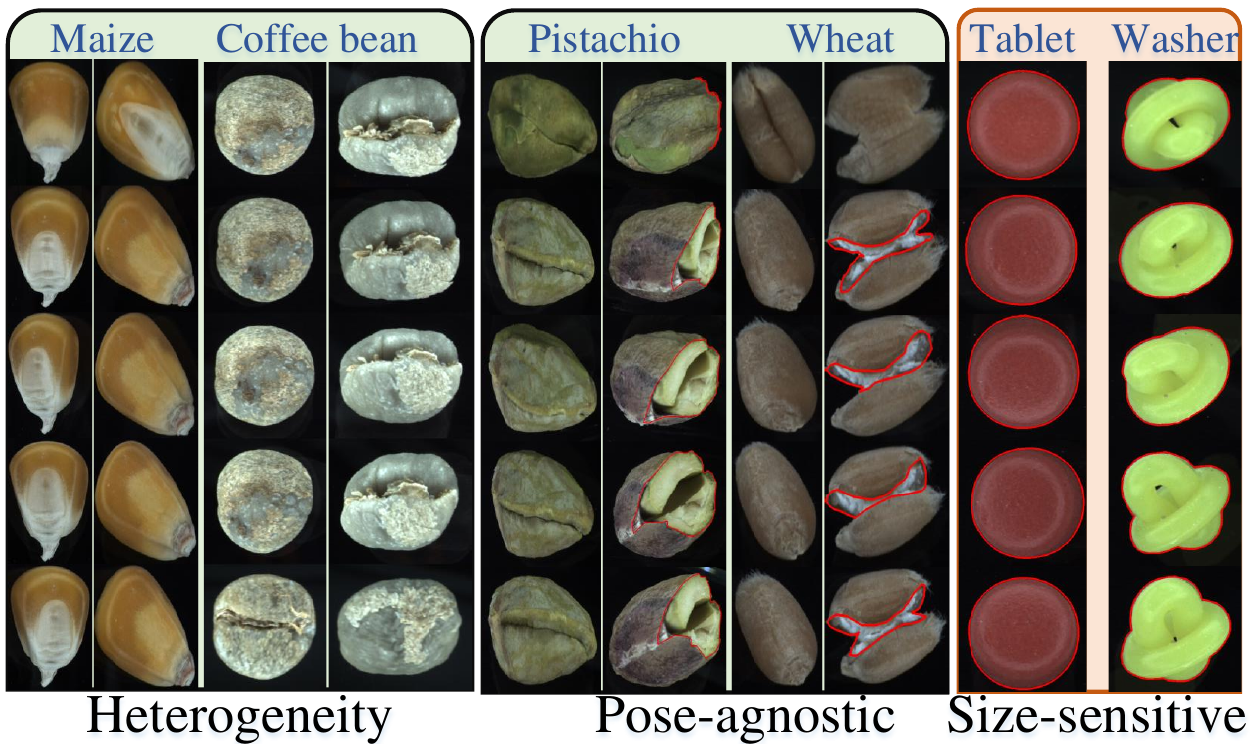}
	\caption{Examples illustrating the image data challenges.} 
	\label{fig:vis_challenge}
\end{figure}

\textbf{Challenges.} Unlike existing datasets that primarily focus on industrial products \cite{mvtec_ad,visa_ad,real_ad}, our dataset presents heterogeneity, pose-agnostic and size-sensitive challenges due to the unique characteristics of tiny objects, as illustrated in Figure ~\ref{fig:vis_challenge}. Specifically, \textit{heterogeneity} refers to the inherent variability observed within normal samples of the same category, as our dataset includes tiny natural products, \eg wheat and soybean kernels, which naturally differ in shape, color, and texture due to growth conditions. \textit{Pose-agnostic} implies that the viewing angles for the same object are agnostic, with precise pose alignment not guaranteed, as controlling the orientation of certain tiny objects is highly complex and costly. \textit{Size-sensitive} highlights the critical importance of size variations for specific objects, especially in fields such as medicine and mechanics, where even minor deviations in size can directly affect dosage accuracy and assembly specifications.

\subsection{Text Annotations and Statistics}

To advance the applicability of anomaly detection datasets in visual-language tasks \cite{radford2021learning,alayrac2022flamingo,li2023blip}, we identified and summarized two common forms of text annotations. First, domain experts often provide detailed descriptions of common anomalies within their field, supplementing specific categories with additional information; we refer to this as \textbf{declarative knowledge}. Second, experts typically use comparative illustrations with normal-normal or normal-anomaly images and pose key questions to guide learning; we refer to this method as \textbf{constructivist learning}.

\begin{figure}[t]
	\centering
	\includegraphics[width=0.48\textwidth]{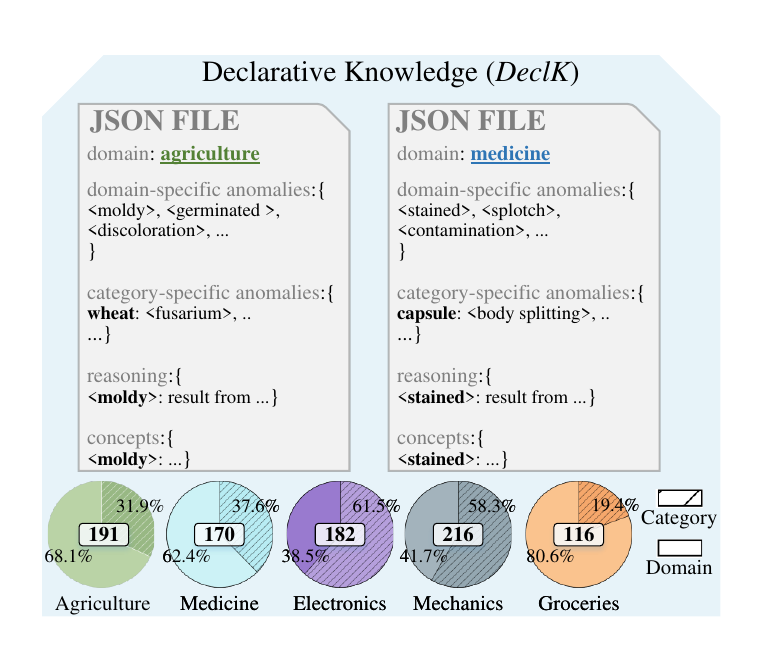}
	\caption{Format and data distribution of Declarative Knowledge.}     
	\label{fig:declk}
\end{figure}

\textbf{Declarative Knowledge \textit{(DeclK)}}. It leverages expert knowledge to produce comprehensive textual descriptions, structured across three levels: explicit instruction, reasoning and concepts corresponding to \textit{$<$what, why, how$>$}, as depicted in Figure \ref{fig:poster}. The motivation is to align with cognitive load theory \cite{apt1988towards}, supporting learners who excel in verbal-linguistic intelligence by providing language-based pathways for understanding.

Specifically, experts codified a specialized vocabulary (noun or adjective) to describe \textit{explicit instruction} for both domain-specific and category-specific anomalies. For example, in the agriculture domain, ``moldy'' represents a domain-specific anomaly, while ``wheat blight'' is a category-specific anomaly for wheat. These terms are further supplemented with detailed explanations of underlying causes, providing \textit{reasoning} and progression of anomalies. To enhance the understanding of \textit{concepts}, anomalies are described through their corresponding visual attributes, \eg color, shape, texture and size. Both \textit{reasoning} and \textit{concepts} annotations are enriched with insights generated by large language models (ChatGPT-4).

\begin{figure}[t]
	\centering
	\includegraphics[width=0.46\textwidth]{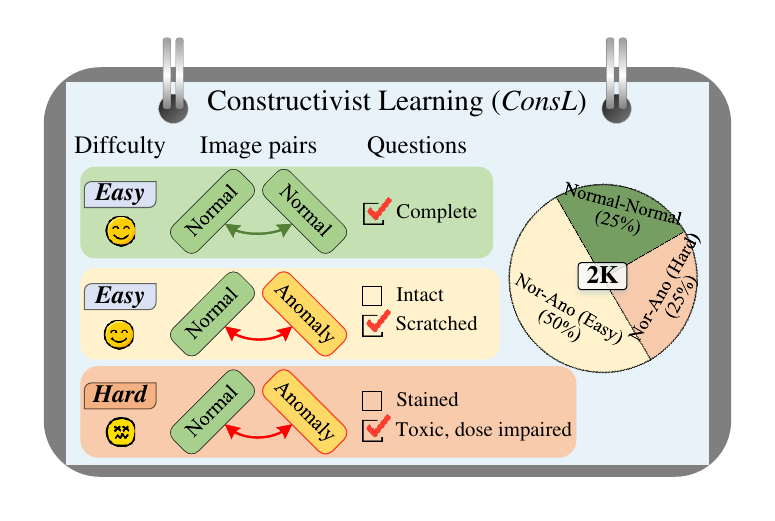}
	\caption{\textbf{Overview of Constructivist Learning}. We provide 2K MCQs with both easy and hard difficulty levels, each containing five options and explanations. } 
	\label{fig:consl}
\end{figure}

We constructed 875 \textit{explicit instructions} for all domains including 391 category-specific words, with corresponding \textit{reasoning} and \textit{concepts} provided in JSON format, as shown in Figure \ref{fig:declk}. All text descriptions and generated descriptions underwent thorough review to ensure accuracy. \textit{DeclK} serves as a structured representation bridging textual and visual domains by creating a detailed textual resource, enhancing the detection and reasoning capabilities of models in visual-language tasks.

\textbf{Constructivist Learning (\textit{ConsL})}. Inspired by learning theory \cite{palincsar1998social} and recent multimodal studies \cite{tong2024mass,tong2024eyes}, we developed \textit{ConsL} which involves creating pair-wise images and formulating corresponding multiple-choice questions (MCQ) based on category-specific anomaly information. The motivation is rooted in dual coding theory, which posits that learners achieve better understanding when information is presented in both visual and verbal formats, actively engaging them in identifying and differentiating defects.

Specifically, we manually constructed an average of over 400 pairs of images and corresponding MCQ within each domain, as shown in Figure ~\ref{fig:consl}. The image pairs consist of both normal-normal and normal-anomalous pairs, with MCQ settings focused on highly relevant and concerned anomalies. Each MCQ is provided with five options and a prompt instructs ``select the correct answer based on the image pair of $<$category name$>$''. For normal-anomalous pairs, we additionally provide a conclusion describing the anomalies in terms of spatial location, shape, size and type. \textit{ConsL} requires the model to identify whether an anomaly exists, describe it, and provide reasoning for its decision. It introduces a structured framework for visual comparison and reasoning, facilitating learning through comparative analysis with text-based questions.

\textbf{Challenges}. Both text data components represent distinct learning approaches, each suited to different cognitive strategies. \textit{DeclK} provides a hierarchical summary of domain knowledge, explaining anomalies from the perspectives of \textit{$<$what, why, how$>$}. However, these descriptions are general depictions of anomalies and are not directly tied to specific images. In contrast, \textit{ConsL} presents pairwise images with specialized questions, focusing on practical anomaly identification without detailed annotations or predefined categories. Both \textit{DeclK} and \textit{ConsL} require strong capabilities from models, aiming to enhance conceptual comprehension and facilitate visual-text reasoning.

%% file: chapters/method.tex
\begin{figure}[b]
	\centering
	\includegraphics[width=0.5\textwidth]{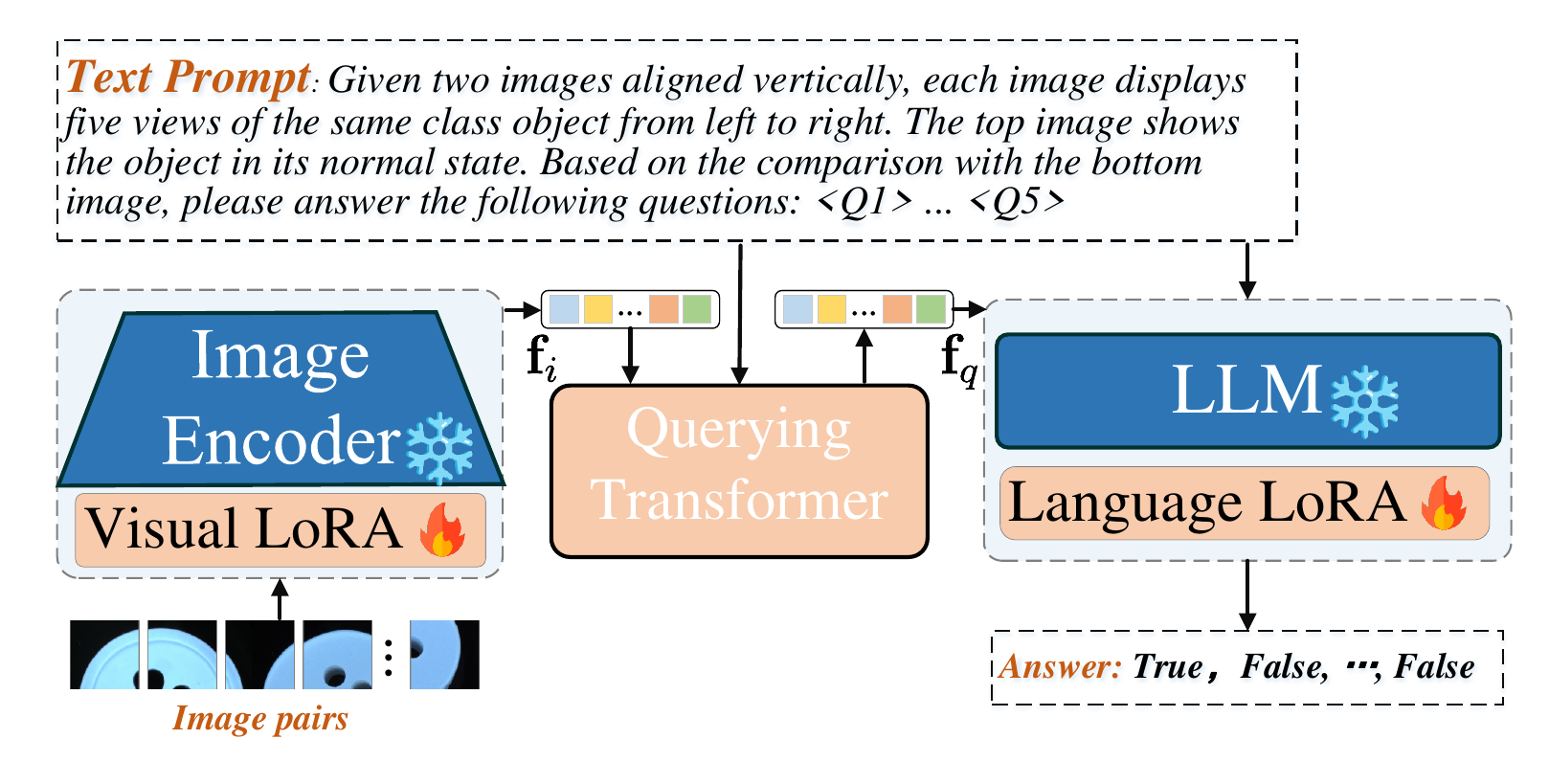}
	\caption{\textbf{Overview of our baseline for \textit{ConsL}.} It consists of an image encoder with a visual LoRA (to extract image features), a Querying Transformer (to bridge the modality gap), and LLM with a language LoRA (for output generation), based on BLIP-2 \cite{li2023blip}.}     
	\label{fig:baseline}
\end{figure}

\begin{figure*}[t]
	\centering
	\includegraphics[width=0.98\textwidth]{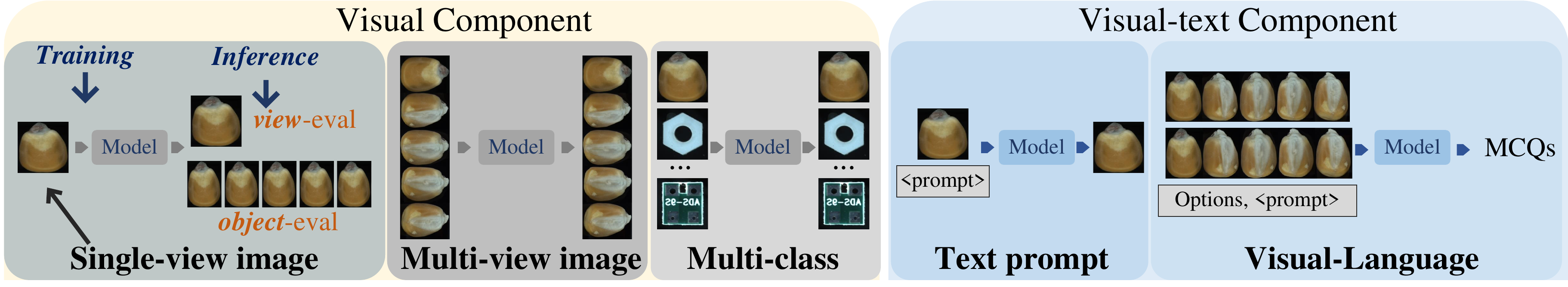}
	\caption{\textbf{Evaluation settings.} As MANTA includes both visual and textural components, with each image having five views, we established five settings according to their training inputs: \textit{Single-view} uses a single view as input for each category; \textit{Multi-view} utilizes all five views of an object for each category; \textit{Multi-class} combines single-view inputs from different categories train a single model; \textit{Text prompt} leverages a single view with text prompts for training; and \textit{Visual-Language} takes a pair of images and prompts to perform MCQ tasks.} 
	\label{fig:setting}
\end{figure*}

\section{Baseline}

For the \textit{ConsL} subset, we explore the application of existing Visual Question Answering and visual-text frameworks \cite{antol2015vqa,radford2021learning}, and fine-tuned a model based on BLIP-2 \cite{li2023blip} model with LoRA \cite{hulora}.

\textbf{Architecture.}
We followed the BLIP-2 \cite{li2023blip} architecture, as illustrated in Figure \ref{fig:baseline}. The model comprises three main modules: an image encoder, a querying transformer, and an LLM. Specifically, the image encoder is a pretrained Vision Transformer used to extract features from input images. The querying transformer consists of two transformer modules that interact with visual features and decode textual output using cross-attention layers, capturing text-related features from visual features and text prompts and thereby bridging the visual and language modality. The LLM includes a pretrained large language model (OPT) \cite{zhang2022opt} used for generating language output.

\textbf{Fine-tuning.}
Both the visual ViT and language OPT models are pretrained on extensive data, so we fix their parameters to reduce computational cost and mitigate catastrophic forgetting of pretrained knowledge. We employ LoRA \cite{hulora} for efficient fine-tuning by using low-rank decomposition matrices.

For each MCQ in \textit{ConsL}, given two images $I_1$ and $I_2$ along with a set of options \(\langle Q_1, \cdots, Q_5 \rangle\), we vertically concatenate the two images and pass them through an image projector to extract image embeddings, which are then fed into the visual ViT to obtain visual features \(\mathbf{f}_i\). We employ cross-entropy loss to classify the input images:
\begin{equation}
    L_{ce} = -\sum_{i=1}^{B}c_i\log(\hat{c_i}) ,
\end{equation}
where \(B\) denotes the batch size, \(c_i\) represents the true class label (\eg button object), and \(\hat{c_i}\) is the predicted class from the input images \([I_1, I_2]\).

We further construct a prompt \(p_t\) based on given options \(\langle Q_1, \cdots, Q_5 \rangle\), which is tokenized to produce the corresponding text features \(\mathbf{f}_t\). The visual features \(\mathbf{f}_i\) and \(\mathbf{f}_t\) are concatenated and fed into the querying transformer to extract the query features \(\mathbf{f}_q\). Finally, \(\mathbf{f}_q\) and \(\mathbf{f}_t\) are combined \([\mathbf{f}_q, \mathbf{f}_t]\) and sent to the language module, which is fine-tuned to produce the final output based on the autoregressive loss.

%% file: chapters/exp.tex
\section{Experiments}

\subsection{Settings and Details}

\textbf{Evaluation Settings}. We established five evaluation settings based on different inputs and training approaches, as shown in Figure \ref{fig:setting}. We first conducted experiments on the visual component to explore advanced methods under both one-for-one (training a single model for each class) and multi-class settings (training a single model for multiple classes). Next, we evaluated the combined visual and textural setup, focusing on few-shot evaluation settings with text prompt learning and visual-language models.

\noindent \textbf{Metrics and Implementation Details}. We reported standard image-level (I-) and pixel-level (P-) Area Under the Receiver Operating Characteristics (AUROC) scores \cite{zavrtanik2021draem,patchcore,pang2021deep} to evaluate model performance in anomaly detection and localization. Complete details on model implementation, additional metrics, and comprehensive experimental results are provided in the \textit{Supp}.

\begin{table*}[t]
\caption{\textbf{Comparisons of \textit{single-view} setting for each class}.  Models are trained using single-view images and reported results in both \textit{view}-eval and \textit{object}-eval (predictions from five views of an object). All results are presented as I-/P-AUROC (\%). We reported only the average results across multiple classes for each domain, with details provided in the \textit{Supp}.}

 \centering
\resizebox{0.98\textwidth}{!}{
   
    \begin{tabular}{lcccccccccc|cc}
    \toprule
    Model('year) & \multicolumn{2}{c}{Agriculture}   & \multicolumn{2}{c}{Medicine}   & \multicolumn{2}{c}{Electronics}  & \multicolumn{2}{c}{Mechanics}  & \multicolumn{2}{c}{Groceries}  & \multicolumn{2}{|c}{Average}  \\
    \cmidrule(lr){2-3} \cmidrule(lr){4-5} \cmidrule(lr){6-7} \cmidrule(lr){8-9} \cmidrule(lr){10-11} \cmidrule(lr){12-13}
     & \textit{view}-eval  & \textit{object}-eval  & \textit{view}-eval  & \textit{object}-eval  & \textit{view}-eval  & \textit{object}-eval & \textit{view}-eval  & \textit{object}-eval  & \textit{view}-eval  & \textit{object}-eval  & \textit{view}-eval  & \textit{object}-eval \\
    \cmidrule(lr){1-13}
    RD'22 \cite{deng2022anomaly}    & 84.7/85.2  & 87.2/85.3  & 86.4/94.0  & 92.3/94.1  & 86.6/93.5  & 89.2/93.4  & 81.0/91.9  & 85.7/91.8  & 71.2/85.3  & 73.5/85.3  & 82.0/90.0  & 85.6/90.0  \\
    PatchCore'22 \cite{patchcore} & 93.0/93.0  & 94.2/93.0  & 96.5/96.7  & 97.4/97.0  & 97.1/98.8  & 97.0/98.8  & 95.8/98.7  & 96.4/98.7  & 86.2/91.5  & 90.8/91.5  & 93.7/95.7  & 95.2/95.8  \\
    CDO'23 \cite{cao2023collaborative}   & 90.0/90.6  & 91.3/90.4  & 95.2/95.4  & 95.2/95.5  & 94.1/98.6  & 91.9/98.6  & 91.1/98.4  & 90.3/98.4  & 85.4/90.4  & 87.7/89.3  & 91.2/94.7  & 91.3/94.4  \\
    DMAD'23 \cite{liu2023diversity}  & 84.5/84.9  & 86.9/85.2  & 84.6/93.3  & 91.2/91.2  & 87.4/92.1  & 90.1/92.2  & 80.2/90.3  & 85.3/90.4  & 75.1/87.5  & 77.0/87.4  & 82.6/89.6  & 86.1/89.7  \\
    SimpleNet'23 \cite{liu2023simplenet} & 86.3/83.3  & 91.2/84.3  & 91.1/87.2  & 94.7/88.3  & 89.6/90.3  & 92.5/91.2  & 86.4/88.7  & 89.6/88.5  & 82.2/82.5  & 86.5/84.9  & 87.1/86.4  & 90.9/87.5  \\
    \bottomrule    
    \end{tabular}%
}
\label{tab:one4one-single}
\end{table*}

\subsection{Results on Visual Component}

\textbf{The \textit{single-view} setting}. 
Considering that existing methods are typically trained using single-view data, including other multi-view datasets (\eg Real-IAD \cite{real_ad}), we evaluated single-view data and conducted nearly 200 experiments to compare several advanced methods: RD \cite{deng2022anomaly}, DMAD \cite{liu2023diversity}, PatchCore \cite{patchcore}, CDO \cite{cao2023collaborative}, and SimpleNet \cite{liu2023simplenet}.
Results were reported using two evaluation strategies: \textit{view}-eval, where each view is treated as an individual test sample, and \textit{object}-eval where five views of an object are considered a single test sample. 

As shown in Table~\ref{tab:one4one-single}, all models achieved performance below 97\% of I-AUROC across all domains, with the agriculture and groceries averaging below 90\% of I-AUROC. This highlights the challenging nature of our dataset compared to other datasets (\eg MVTec \cite{mvtec_ad}) with 100\% of I-AUROC, as the heterogeneity and pose-agnostic challenges in our data require models to effectively capture complex normal patterns. Additionally, we observed that reconstruction-based models performed relatively worse compared to other approaches. we attribute this to the fact that many anomalous samples in tiny objects possess large anomalous regions, which reconstruction-based methods often struggle to address effectively.

We observed that performance for \textit{view}-eval and \textit{object}-eval are similar. \textit{view}-eval is computed separately for each individual view, while \textit{object}-eval aggregates five views of an object before calculation. This grouping effectively averages out variations in model performance across different views, leading to a more stable overall evaluation.

\begin{table}[b]
\caption{\textbf{Comparisons of \textit{multi-view} setting for each class}. Models are trained using multi-view images. We presented only the averaged results (I-/P-AUROC, \%) across multiple classes for each domain, with details provided in the \textit{Supp}.}
 \centering
\resizebox{0.48\textwidth}{!}{
    \begin{tabular}{lccccc|c}
    \toprule
    Model & Agriculture & Medicine & Electronics & Mechanics & Groceries & Average \\
    \cmidrule(lr){1-7}
    RD'22 \cite{deng2022anomaly}    & 86.9/89.5  & 94.3/96.1  & 94.9/98.5  & 96.1/98.7  & 81.1/87.3  & 90.6/94.0  \\
    PatchCore'22 \cite{patchcore} & 94.1/93.0  & 96.8/96.2  & 96.8/98.7  & 96.6/98.6  & 91.0/91.8  & 95.0/95.7  \\
    CDO'23 \cite{cao2023collaborative}   & 90.7/90.4  & 95.4/95.5  & 92.3/98.6  & 91.3/98.4  & 87.8/89.5  & 91.5/94.5  \\
    DMAD'23 \cite{liu2023diversity}  & 88.5/89.7  & 95.1/95.8  & 94.2/98.3  & 93.3/98.3  & 80.0/86.3  & 90.2/93.7  \\
    \bottomrule
    \end{tabular}%
}
\label{tab:one4one-multi}
\end{table}

\textbf{The \textit{multi-view} setting}. 
We conducted over 150 experiments using five-view images of an object as individual training samples, with results reported at the \textit{object}-eval level. Since the size of multi-view inputs is five times larger than single-view images, we adapted several advanced methods, including RD \cite{deng2022anomaly}, Patchcore \cite{patchcore}, CDO \cite{cao2023collaborative}, and DMAD \cite{liu2023diversity}, to accommodate this configuration due to memory constraints.

As shown in Table~\ref{tab:one4one-multi}, all models demonstrated promising performance across the five domains, with an average of I-AUROC of 91\% and P-AUROC of 94\%. Compared to \textit{single}-view inputs, we attribute this improved performance to the \textit{multi-view} images providing multiple perspectives that share critical features of the object. However, considering the computational cost of using multi-view inputs, our subsequent experiments primarily focused on \textit{single-view} data.

\textbf{The \textit{multi-class} setting}. 
We also evaluated the popular multi-class setting, where all classes within each domain are mixed for the single-view training. We evaluated recent methods, including UniAD \cite{uniAD}, CRAD \cite{lee2024crad} and HGAD \cite{HGAD}, as shown in Table~\ref{tab:one4all}. These models exhibited performance below I-AUROC of 90\% and P-AUROC of 91\% across the five domains, representing a decline compared to the \textit{single-view} and \textit{multi-view} settings. We attribute this to the variability among objects, even within the same domain. For example, in the agriculture domain, wheat and maize exhibit notable differences, while in electronics, the appearances of Type-C connectors and LEDs differ significantly. This variability imposes high demands on multi-class models to generalize and learn normal patterns across different classes simultaneously.

\begin{table}[t]
\caption{\textbf{Comparisons of \textit{multi-class} setting}. Models are trained using single-view mixed data across all categories within each domain. Results are presented as I-/P-AUROC (\%).}
 \centering
\resizebox{0.48\textwidth}{!}{
    \begin{tabular}{lccccc|c}
    \toprule
    Model('year) & Agriculture & Medicine & Electronics & Mechanics & Groceries & Average \\
    \cmidrule(lr){1-7} 
    UniAD'23 \cite{uniAD} & 70.5/81.9  & 91.3/93.3  & 93.6/97.7  & 90.2/97.2  & 66.7/76.8  & 82.4/89.4  \\
    CRAD'24 \cite{lee2024crad}  & 85.2/86.4  & 94.2/94.6  & 94.5/96.5  & 92.9/96.9  & 76.1/83.5  & 88.6/91.6  \\
    HGAD'24 \cite{HGAD}  & 85.5/88.6 & 92.6/93.7  & 91.0/94.3  & 83.5/94.8  & 74.8/71.7  & 85.5/88.6 \\
    \bottomrule    
    \end{tabular}%
}
\label{tab:one4all}
\end{table}

\subsection{Results on Text Component}

As MANTA includes two text subsets: \textit{DeclK} and \textit{ConsL}, we explored two types of visual-text settings: \textbf{Text prompt}, which uses text prompts from \textit{DeclK} combined with images to train a model for anomaly detection; and \textbf{Visual-Language}, which leverages images, options, and conclusions from \textit{ConsL} to fine-tune our baseline model.

\begin{figure}[b]
	\centering
	\includegraphics[width=0.46\textwidth]{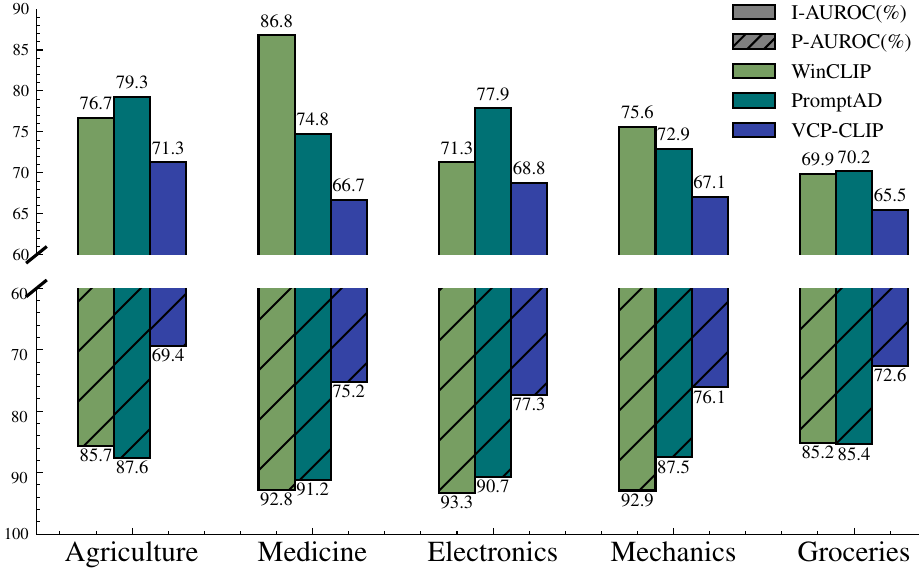}
	\caption{\textbf{Comparison of \textit{text prompt} setting in one-shot learning}. Models are trained using text data from \textit{DeclK}. } 
	\label{fig:abl-prompt}
\end{figure}

\begin{figure*}[t]
	\centering
	\includegraphics[width=0.93\textwidth]{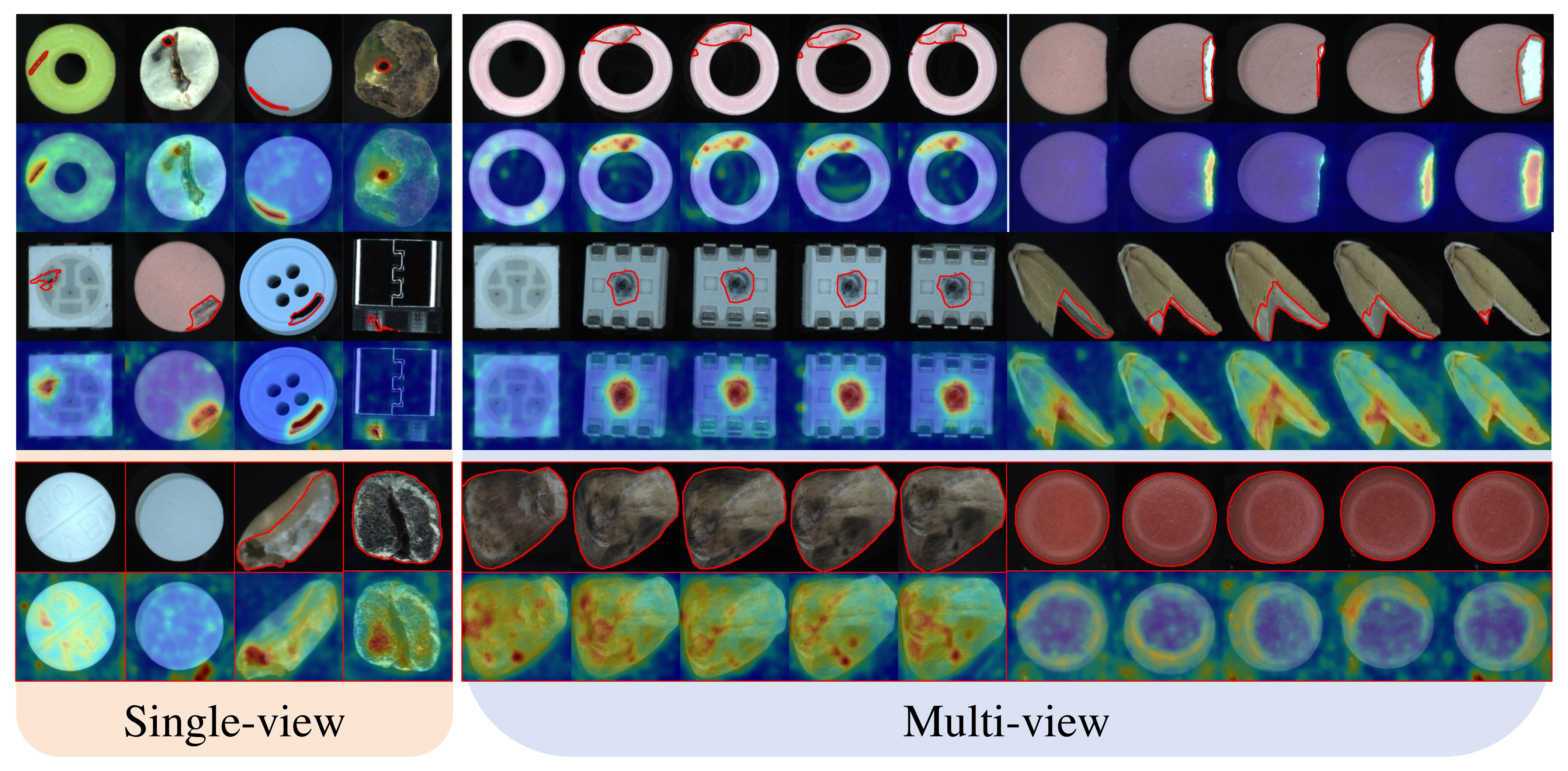}
	\caption{\textbf{Visualization of \textit{single-view} and \textit{multi-view}.} We presented both successful and failure cases to qualitatively illustrate the challenges posed by our MANTA dataset. }     
	\label{fig:vis}
\end{figure*}

\textbf{The \textit{text-prompt} setting}. 
We explored our \textit{DeclK} subset using advanced text prompt AD methods, including WinCLIP \cite{jeong2023winclip}, PromptAD \cite{li2024promptad} and VCP-CLIP\cite{qu2024vcp}. Due to computational resource constraints, we focused on one-shot learning and constructed a small-scale test set of 1000 images, with equal numbers of normal and anomalous samples per domain. For the text prompt, we randomly selected explicit instructions from \textit{DeclK} (formatted as ``$\left<instruction\right>$ of $\left<category\right>$'') along with visual concepts as negative samples, while we randomly sampled other explicit instructions, using the format ``without $\left<instruction\right>$'' as positive samples.

As shown in Figure \ref{fig:abl-prompt}, the three methods achieved moderate performance under the one-shot learning setting, with WinCLIP surpassing an average of 75\% in I-AUROC and 90\% in P-AUROC across five domains. This indicates that our \textit{DeclK} subset can serve as effective supervision, bridging the gap to extract visual-text features from foundation models \cite{radford2021learning}, enhancing anomaly detection capabilities.

\begin{table}[b]
\caption{\textbf{Performance of \textit{visual-language} setting}. Our baselines are tested in zero-shot and one-shot settings. Results are presented as \textit{option}/\textit{question} (\%) accuracy. N-A(hard) denotes Normal-Anomaly image pairs with a hard level of difficulty.}
 \centering
\resizebox{0.38\textwidth}{!}{
    \begin{tabular}{ccccc}
    \toprule
    Settings & N-N(easy) & N-A(easy) & N-A(hard)  & Average \\
    \cmidrule(lr){1-5} 
    zero-shot & 44.8/3.5 & 51.8/2.5 & 38.7/1.5  & 45.1/2.5 \\
    few-shot &  \textbf{53.8}/\textbf{6.9} & \textbf{62.0} /\textbf{5.5} & \textbf{50.7}/\textbf{4.6}  &\textbf{55.2}/\textbf{5.7} \\
    \bottomrule
    \end{tabular}%
}
\label{tab:mcq}
\end{table}

\textbf{The \textit{visual-language} setting}. 
We evaluated our baseline model on \textit{ConsL} in both zero-shot and few-shot learning. We randomly selected 50\% of the data from each domain with balanced difficulty levels for training, while the remaining 50\% was used for testing. We reported two metrics: \textit{option} accuracy, where each option is scored individually, and \textit{question} accuracy, where an entire MCQ is considered correct only if all options are correctly answered.
We vertically concatenated two images and provided the text prompt: ``\textit{Given two images aligned vertically, each image displays five views of the same class object from left to right. The top image shows the object in its normal state. Based on the comparison with the bottom image, please answer the following questions: $\left<Q1\right>, \cdot, \left<Q5\right>$..}''.

As shown in Table \ref{tab:mcq}, compared to the zero-shot learning, the model using few-shot learning achieved an average of 50.2\% and 5.7\% in \textit{option} and \textit{question} accuracy respectively, but still exhibited unsatisfactory performance. We attribute this to the domain gap between our images and the pretrained data of the baseline's visual ViT. Additionally, in terms of \textit{question} accuracy, especially for hard MCQs, the models need to accurately identify anomalies and infer the correct options based on the nature of the anomaly and the object's category, which demands strong visual understanding and reasoning capabilities.

\subsection{Visualization}

We presented qualitative results for the single-view and multi-view settings, as illustrated in Figure \ref{fig:vis}. In the \textit{single-view} setting, the model effectively localized smaller anomalous regions, such as insect-attacked holes, and mold spots on grains. However, larger anomalies often led to missed areas, such as complete impurities or larger broken regions. In the \textit{multi-view} setting, the model demonstrated good detection of small anomalous regions and maintained consistency across multiple views. However, due to the larger image size, more localization errors were observed.

%% file: chapters/conc.tex
\section{Conclusion}

In this paper, we introduce MANTA, a comprehensive visual-text anomaly detection dataset for tiny objects spanning five domains. With over 137K multi-view images and two text subsets, our dataset presents unique challenges and diverse settings for anomaly detection. Extensive experiments highlighted the complexities of MANTA, paving the way for advancements in visual-language applications.

\textbf{Limitations}. MANTA focuses on tiny objects, which presents significant differences compared to other conventional objects. Although we utilized five cameras, certain areas may remain partially uncovered (\eg hexagonal sides). Furthermore, due to cost constraints, the size of our text data subset is limited.